\pdfoutput=1

\documentclass[11pt]{article}

\usepackage{ACL2023}

\usepackage{times}
\usepackage{tabularx}
\usepackage{lipsum}
\usepackage{hyperref}
\usepackage{float}
\usepackage{xcolor}
\usepackage{latexsym}
\usepackage{amsmath}
\usepackage{amsfonts}
\usepackage{dsfont}
\usepackage{graphicx}
\usepackage{caption}
\usepackage{subcaption}
\usepackage{multirow}
\usepackage{booktabs}
\usepackage{algorithm}
\usepackage{algpseudocode}
\usepackage{colortbl}
\definecolor{lightgray}{gray}{0.9}
\definecolor{lightgreen}{rgb}{0.9, 1, 0.9}
\usepackage{xcolor}
\usepackage[utf8]{inputenc}

\usepackage{microtype}

\usepackage{inconsolata}

\title{Knowledge-to-SQL: Enhancing SQL Generation with Data Expert LLM}

\author{
Zijin Hong\textsuperscript{1}, Zheng Yuan\textsuperscript{2}, Hao Chen\textsuperscript{2}, Qinggang Zhang\textsuperscript{2} \\
{\bf Feiran Huang\textsuperscript{1}\footnotemark[2]  , Xiao Huang\textsuperscript{2}}\\ 
\textsuperscript{1}Jinan University \\
\textsuperscript{2}The Hong Kong Polytechnic University\\ 
\texttt{hongzijin@stu2020.jnu.edu.cn} \\
\texttt{\{yzheng.yuan,qinggangg.zhang\}@connect.polyu.hk} \\
\texttt{sundaychenhao@gmail.com}; \texttt{huangfr@jnu.edu.cn}; \texttt{xiaohuang@comp.polyu.edu.hk}
}

\begin{document}
\maketitle
\renewcommand{\thefootnote}{\fnsymbol{footnote}} 
\footnotetext[2]{Corresponding author}
\renewcommand{\thefootnote}{\arabic{footnote}}

\begin{abstract}
Generating accurate SQL queries for user questions (text-to-SQL) has been a long-standing challenge since it requires a deep understanding of both the user's question and the corresponding database schema in order to retrieve the desired content accurately.
Existing methods rely on the comprehensive capability of large language models (LLMs) to generate the SQL.
However, some necessary knowledge is not explicitly included in the database schema and user question or has been learned by LLMs. 
Thus, the generated SQL of the knowledge-insufficient questions may be inaccurate, negatively influencing the text-to-SQL models' performance and robustness. 
To address this challenge, we propose the \textbf{Knowledge-to-SQL} framework, which employs tailored \textbf{Data Expert LLM (DELLM)} to provide helpful knowledge for all text-to-SQL models. 
Specifically, we introduce the detailed implementation of DELLM regarding table reading and the basic fine-tuning process. 
We further propose a \textit{\underline{P}reference \underline{L}earning via \underline{D}atabase \underline{F}eedback (PLDBF)} strategy, refining the DELLM to generate more helpful knowledge for LLMs. 
Extensive experiments verify that DELLM can enhance the state-of-the-art approaches for text-to-SQL tasks. 
The corresponding code of DELLM is released for further research\footnote{\href{https://github.com/Rcrossmeister/Knowledge-to-SQL}{https://github.com/Rcrossmeister/Knowledge-to-SQL}}. 
\end{abstract}

\section{Introduction}

Generating SQL based on user questions (text-to-SQL) is currently one of the leading applications for large language models (LLMs). 
The most straightforward approach is to input the user questions and database schema into the LLMs and rely on their capability of natural language understanding to generate the SQL~\cite{dou2022towards,liu2023comprehensive,pourreza2023dinsql}. 
However, in real-world applications, user queries and database structure may contain customized or specialized knowledge~\cite{yu2018Spider,li2023BIRD}, including arithmetic reasoning, domain knowledge, synonym explanation, etc. 
Since the knowledge is not explicit in either the user question or the database schema~\cite{dou2022towards,gan2021Spider-DK}. 
In such cases, the intuitive approach may result in inaccurate or un-executable SQL unless a human expert provides the necessary helpful knowledge (so-called \textbf{``expert knowledge''}) to the LLMs to bridge the knowledge gap with database content~\cite{li2023BIRD,gan2021Spider-DK}.
Given this challenge, it is valuable to develop a non-human data expert system that can automatically generate the required expert knowledge to assist SQL generation. This would significantly enhance the performance and robustness of text-to-SQL implementations.

\begin{figure}[t]
    \centering
    \includegraphics[width=0.45\textwidth]{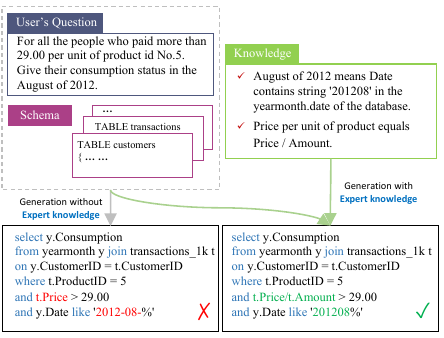}
    \caption{A sketch map illustrating the significance of incorporating expert knowledge in the text-to-SQL implementation. In the given example, the generation without expert knowledge makes mistakes in arithmetic reasoning and data conditions. Expert knowledge bridges the knowledge gap between the LLMs and the database, which assists the LLMs in generating accurate SQL.}
    \label{fig:intro}
\end{figure}

Existing methods primarily focus on fully exploring the comprehensibility of the pre-trained language models (PLMs). 
As the forerunner, T5-based methods~\cite{scholak2021picard,li2023graphix,rai2023improving} initially attempted to train PLMs from scratch to generate SQL based on user questions and database schema. 
Following the popularity of proprietary LLMs, particularly ChatGPT and GPT-4~\cite{openai2023gpt}, DIN-SQL~\cite{pourreza2023dinsql} utilizes LLMs to decompose the process of SQL generation into several sub-tasks. 
It uses LLMs to first accomplish these sub-tasks and then aggregates the results to generate the final SQL. 
Similarly, DAIL-SQL~\cite{gao2023text} designed a strategy for sampling few-shot instances, then using few-shot prompting for proprietary LLMs to accomplish the generation of the required SQL. 
Most recently, MAC-SQL~\cite{wang2023mac} assigns different roles to three LLM agents, specifically selector, decomposer, and refiner, then uses these agents to make different contributions to the text-to-SQL process.

However, recent methods mainly focus on designing more sophisticated structures in order to improve text-to-SQL performance. 
Although achieving promising performance, there is a lack of emphasis on the necessity of expert knowledge~\cite{li2023BIRD}. 
As depicted on the left of Figure~\ref{fig:intro}, relying solely on the user's question and the database schema as input without expert knowledge may cause inaccurate conditions in the generated SQL. 
For example, the incorrect condition ``y.Date like "2012-08-\%"'' could potentially result in the SQL returning empty data. 
In contrast, on the right of the figure, when provided with accurate expert knowledge for assistance, the LLMs are able to rectify the condition and generate valid SQL.

Nevertheless, generating expert knowledge from the question and schema faces the following challenges: 
\textbf{1. Question \& Database Specialization}: The generated expert knowledge should be specialized for the given question and database.
It is challenging for the data expert to understand the question and database and provide helpful knowledge. 
\textbf{2. Table Content Awareness}~\cite{cheng2023binding}: The data expert should be able to read the content of the table in order to determine whether it is necessary to provide detailed content examples in the knowledge. 
\textbf{3. Performance Enhancement}~\cite{li2023BIRD}: The generated knowledge should be helpful for the text-to-SQL models. 
It is challenging to ensure that expert knowledge can contribute to more accurate SQL generation.

By addressing the aforementioned challenges, we present a detailed design of \textbf{Knowledge-to-SQL} framework for enhancing the SQL generation of LLMs. 
Specifically, we propose a well-designed \textbf{Data Expert Large Language Model (DELLM)}, which comprises a table reading module and a knowledge-oriented supervised fine-tuning process. 
Furthermore, we introduce \textit{\underline{P}reference \underline{L}earning via \underline{D}atabase \underline{F}eedback (PLDBF)} to further refine the helpfulness of the generated expert knowledge for LLM-based text-to-SQL~\cite{christiano2017rlhf,rafailo2023direct,hong2024towards}. 
Specifically, PLDBF provides preferences to DELLM based on two tailored criteria: 1) the extent to which the generated knowledge aids in retrieving more accurate content, and 2) the degree to which the generated knowledge assists LLM in producing more precise SQL queries.
In summary, our contributions can be listed as follows:

\begin{itemize}
    \item We highlight the significance of expert knowledge and present the knowledge-to-SQL framework for improving SQL generation.
    \item We introduce a well-designed Data Expert Large Language Model (DELLM), along with customized structure, fine-tuning technique, and preference-tuning training strategies.
    \item We release the training and evaluation code of DELLM as open source for future research.
    \item We validate the effectiveness of our approach on the BIRD and Spider datasets, demonstrating that DELLM can generally enhance the performance of common LLM-based text-to-SQL implementations.
\end{itemize}

\section{Related Work}
\subsection{LLMs for Text-to-SQL}
The text-to-SQL task focuses on translating natural language questions into SQL queries. 
Recent advances in this field have shown a growing interest in using large language models (LLMs) paired with prompt engineering.
Chain of Thought (CoT) prompting~\cite{wei2022cot}, an effective prompt engineering technique, has found considerable utility in the text-to-SQL domain. 
Numerous researchers conducted empirical studies on prompt organization for text-to-SQL using CoT~\cite{rajkumar2022evaluating,gao2023text,chang2023prompt,li2023BIRD,zhang2023act}. 
In DAIL-SQL~\cite{gao2023text}, the authors introduced a novel approach for selecting pertinent few-shot examples, considering both similarities in the user's question and SQL query. 
Moreover, ACT-SQL~\cite{zhang2023act} employed a cost-efficient method to automatically generate CoT instances, alleviating the need for manual prompt creation. 
For enhancing the reasoning capabilities of LLMs, works like C3 ~\cite{dong2023c3} and DIN-SQL~\cite{pourreza2023dinsql} have proposed a paradigm for decomposing main tasks into multiple sub-tasks, which have been specifically designed for zero-shot and few-shot text-to-SQL tasks respectively. 
Most recently, MAC-SQL~\cite{wang2023mac} assigns multiple LLM agents with different roles for the text-to-SQL process. 
Additionally, to assist text-to-SQL with expert knowledge, authors in~\cite{li2023BIRD} engaged human experts to annotate helpful knowledge for each text-to-SQL instance. 
However, this mode of human-led annotation proves to be highly labor-consuming for large-scale tasks. 
In this paper, we focus on exploring automated expert knowledge generation for assisting text-to-SQL using LLMs.

\subsection{Prompt Engineering in Text-to-SQL}
Previous studies have highlighted the importance of prompt engineering in optimizing the performance of LLMs~\cite{radford2019unsupervisedmultitask, liu2023prompt}. 
The effective prompt design significantly improves the efficiency and quality of LLM outputs~\cite{wei2022zeroshotlearners}. 
Research initiatives like CoT~\cite{wei2022cot} and retrieval augmented generation (RAG)~\cite{lewis2020rag} integrate contextual information to enhance LLMs' understanding of natural language reasoning. 
In the text-to-SQL task, recent investigations~\cite{rajkumar2022evaluating} have refined the prompts for LLMs, resulting in high uniformity input formats. 
These prompts typically include user questions, database schema, and task instructions. 
However, comprehending complex input schema and correlating them with user questions poses a significant challenge for LLMs, stemming from diverse aspects of prompt design. 
Leading approaches~\cite{rajkumar2022evaluating,pourreza2023dinsql} have incorporated techniques such as few-shot sampling~\cite{gao2023text} and CoT~\cite{li2023BIRD,wang2023mac} guidelines to address this challenge. 
Our proposed framework can also be viewed as a particular prompt engineering, with the purpose of generating expert knowledge as input prompts to assist LLMs' understanding of text-to-SQL.

\section{Proposed Method}
\begin{figure*}[t]
    \centering
    \includegraphics[width=1\linewidth]{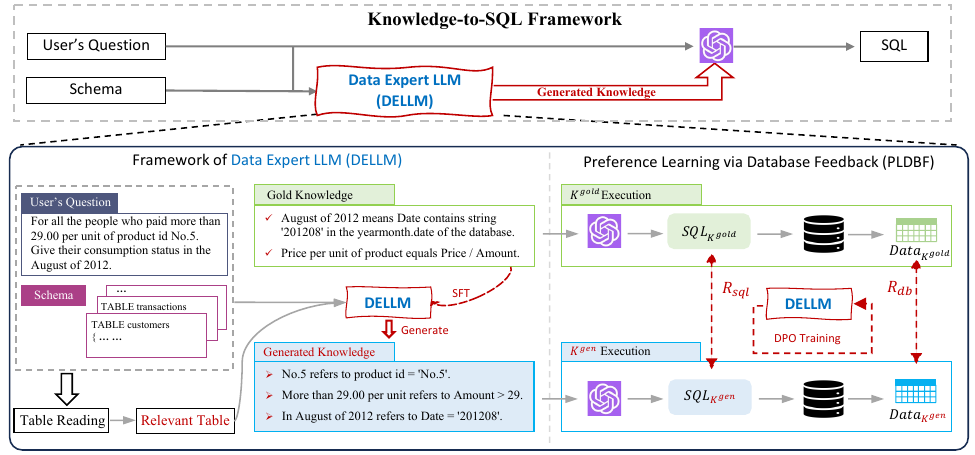}
    \caption{The overview of our approach. The upper is the overall knowledge-to-SQL framework. The details of DELLM are presented at the bottom. On the left side, we have the framework of DELLM, including supervised fine-tuning (SFT) and table reading. On the right side, we introduce preference learning via database feedback (PLDBF), which is employed to further refine the performance of DELLM.}
    \label{fig:framework}
\end{figure*}
We adhere to the two criteria mentioned to guide knowledge generation: 1) Enhancing accurate database execution. 2) Improving precise SQL generation. 
Our proposed method aims to develop a system called DELLM for expert knowledge generation that satisfies the above criteria. 
As depicted in Figure~\ref{fig:framework}, our framework consists of the following three primary modules:
\begin{itemize}
    \item The supervised fine-tuned (SFT) model: It takes the user question, database schema, and relevant tables from the database as input and converts them into generated knowledge.
    \item The preference learning (PL) framework: It refines the model by aligning the feedback from SQL query executions on the database with the contributions from ground-truth SQL.
    \item An off-the-shelf text-to-SQL model: It predicts the SQL by inputting the user question, database schema, and generated knowledge.
\end{itemize}
\subsection{Supervised Fine-Tuning of DELLM}\label{sec:sft}
This module aims to generate expert knowledge based on the user question and the database schema. 
Assuming the user question and the corresponding database schema are $Q$ and $\mathcal{S}$ respectively; the goal is to match the relevant sub-tables $\mathcal{T}_{\alpha}$ and generate knowledge $K^{gen}$ with the above inputs.
\paragraph{Table Reading.}
In our study, we incorporate the task of table reading to generate expert knowledge. 
When dealing with large databases, inputting the complete database tables poses a challenge regarding input length limitation and redundancy.
To address this, we utilize semantic techniques to match the relevant table. 
This allows us to extract the most pertinent table for the given question as the input prompt for the subsequent SFT model. 
Let $\mathcal{S}$ represent the database schema, defined as:
\begin{equation}
    \mathcal{S} = \{\mathcal{T}_1(c_{11},..., c_{1m_{1}}), ... ,\mathcal{T}_i(c_{i1},..., c_{im_{i}}), ...\},
\end{equation}
where $\mathcal{T}_i$ denotes the $i^{th}$ table in the database, and $c_{ij}$ represents the $j^{th}$ column in table $\mathcal{T}_i$, $m_{i}$ denotes the number of column included in $\mathcal{T}_i$. 
Given a schema with $n$ tables, we can denote the collection of all table's columns as $\mathcal{C} = \{\{c_{11},..., c_{1m_{1}}\},..., \{c_{n1},..., c_{nm_{n}}\}\}$. For each column $c_{ij} \in \mathcal{C}$, we can obtain a sub-collection of relevant columns denoted as $\mathcal{C}_{\alpha}$:
\begin{equation}
    \mathcal{C}_{\alpha} = \{c_{ij} \mid \operatorname{sim}(Q, c_{ij}) > \alpha\},
\end{equation}
where $\operatorname{sim}(\cdot)$ denotes the semantic similarity calculator, and $\operatorname{sim}(Q, c_{ij}) > \alpha$ represents the similarity between column $c_{ij}$ and query $Q$ exceeding the threshold $\alpha$. The relevant sub-tables can be obtained by:
\begin{equation}
    \mathcal{T}_{\alpha} = \{\mathcal{T}_{i}(c_{ij}) \mid c_{ij} \in \mathcal{C}_{\alpha}\}.
\end{equation}

\paragraph{Supervised Fine-tuning.}
The knowledge generation process can be represented as follows:
\begin{equation}\label{eq:input}
    K^{gen} = \pi(Q, \mathcal{S}, \mathcal{T}_{\alpha}),
\end{equation}
$\pi(\cdot)$ represents a text generation model we utilize as a backbone model for the SFT process. As defined earlier, the model is required to generate the knowledge $K^{gen}$. The objective function of SFT can be defined as:
\begin{equation}\label{eq:SFT}
\begin{aligned}
    \mathcal{L}_{SFT} &= -\operatorname{log}\operatorname{Pr}(K^{gold} \mid Q, \mathcal{S}, \mathcal{T}_{\alpha}) \\
    & = -\sum K^{gold} \log(K^{gen}),
\end{aligned}
\end{equation}
where $K^{gold}$ denotes the gold (ground-truth) knowledge annotated by human experts. The SFT process aims to minimize the cross-entropy loss in Eq.~\ref{eq:SFT} defined above. Consequently, we can obtain a fine-tuned model $\pi^{SFT}$, which possesses the ability to generate knowledge in a preliminary manner.

\subsection{Feedback From Database Execution and Ground-truth SQL Contribution}
To enable our model to generate helpful knowledge for accurate database execution and contribute effectively to SQL queries, we employ the preference learning framework. 
This framework refines the model's capabilities by aligning feedback from database executions and the contributions of ground-truth SQL.
\paragraph{Feedback From Database Execution.}
In Section~\ref{sec:sft}, we obtained a model $\pi^\text{SFT}$ via SFT. 
We utilize the knowledge generated by this model to assist an off-the-shelf text-to-SQL model $g(\cdot)$ in SQL generation.
\begin{equation}
    \hat{Y}^{gen} = g(\mathcal{S}, Q, K^{gen}),
\end{equation}
where $\hat{Y}^{gen}$ represents the SQL query generated using the knowledge $K^{gen}$. 
For every instance in the training set, we obtain a predicted SQL; the collection of these SQL queries is denoted by $\mathcal{\hat{Y}}^{gen}$. 
Similarly, we obtain a predicted SQL collection $\mathcal{\hat{Y}}^{gold}$ using the manually annotated ground-truth knowledge $K^{gold}$. 
To interact with the database execution, we execute the SQL collections $\mathcal{\hat{Y}}^{gen}$ and $\mathcal{\hat{Y}}^{gold}$, respectively, to obtain the result sets $\hat{V}^{gen}$ and $\hat{V}^{gold}$. An indicator function is defined to evaluate the execution results. We have:
\begin{equation}
    \mathds{1}_{db}(V, V') =
    \begin{cases} 
    1, & V = V' \\
    0, & V \neq V'
    \end{cases},
\end{equation}
then, we utilize this indicator function as the condition for annotating the preference knowledge pairs with the feedback of database execution:
\begin{align}\label{eq:dbf}
&   \mathcal{P}^{db}_{\{K_{w}, K_{l}\}} = \\ 
&  \{K^{gold}_i,K^{gen}_i \mid \mathds{1}_{db}(\hat{V}^{gold}_{i}, \hat{V}^{gen}_{i}) = 0 \}. \nonumber
\end{align}
This feedback requires the generated knowledge to align with database execution.

\paragraph{Feedback From Ground-truth SQL Contribution.}
\begin{table}[!t]
\centering
\scalebox{0.9}{
\begin{tabularx}{\linewidth}{X}
\toprule
\textbf{Ground-Truth SQL:} \\
\textbf{SELECT} `Free Meal Count (Ages 5-17)' / `Enrollment (Ages 5-17)' \textbf{FROM} from \textbf{WHERE} `Educational Option Type' = `Continuation School' \textbf{AND} \textcolor{orange}{`Free Meal Count (Ages 5-17)' / `Enrollment (Ages 5-17)'} \textbf{IS} \textbf{NOT} \textbf{NULL} \textbf{ORDER} \textbf{BY} `Free Meal Count (Ages 5-17)' / `Enrollment (Ages 5-17)' \textbf{ASC} \textbf{LIMIT} 3
\\
\midrule
\textbf{Contributing Knowledge:} \\
Eligible free rates for students aged 5-17 = \textcolor{blue}{`Free Meal Count (Ages 5-17)' / `Enrollment (Ages 5-17)'}.
\\ 
\midrule
\textbf{Non-contributing Knowledge:} \\
Continuation schools refer to \textcolor{red}{EdOpsCode = `C'}, lowest three eligible free rate refer to \textcolor{red}{MIN(`Percent (\%) Eligible Free (Ages 5-17)')}.\\ 
\bottomrule
\end{tabularx}}
\caption{Example of ground-truth SQL contribution. We highlight the key content in SQL and the corresponding sub-knowledge included in the contributing and non-contributing knowledge.}
\label{tab:conexample}
\label{tab:}
\end{table}

Knowledge plays a decisive role in text-to-SQL that the final predicted SQL will include sub-knowledge content introduced by the given knowledge. 
An example is shown in Table~\ref{tab:conexample}.
However, the incorrect sub-knowledge may lead to a misprediction in generating SQL queries. 
A contributing knowledge indicates that every included sub-knowledge is helpful for the predicted SQL, the corresponding content should be valid. 
Thus, we focus on studying the contribution of generated knowledge to the ground-truth SQL.  

Noting that a generated knowledge $K$ comprises several sub-knowledge $k_1, k_2, ...$, can be represented as $K = \{k_1, k_2, ...\}$. We introduce another indicator function:
\begin{equation}
    \mathds{1}_{sql}(K, Y) =
    \begin{cases} 
    1, & k \in Y,\, \forall k \in K\\
    0, & k \notin Y,\, \exists k \in K
    \end{cases},
\end{equation}
where $Y$ is the ground-truth SQL. We check whether every sub-knowledge is contained by $Y$ to judge whether the knowledge contributes. 
Then, we collect the preference knowledge pairs based on the feedback of ground-truth SQL contribution, formulated as:
\begin{align}\label{eq:sqlf}
\mathcal{P}^{sql}_{\{K_{w}, K_{l}\}} = \{ & K^{gold}_j,K^{gen}_j \mid \mathds{1}_{sql}(K^{gold}_{j}, Y) = 1,\\
& \mathds{1}_{sql}(K^{gen}_{j}, Y) = 0 \}. \nonumber
\end{align}
This feedback necessitates the generated knowledge to become correctly SQL contributing and reduce the redundancy in the knowledge generation process.

By Eq.~\ref{eq:dbf} and \ref{eq:sqlf}, we obtain two preference knowledge set $\mathcal{P}^{db}$ and $\mathcal{P}^{sql}$. 
To get the final preference learning dataset $\mathcal{D}$, we combine corresponding input of $K_{l}$ according to Eq.~\ref{eq:input}, denoted by $I_{l}$ and the preference pair $(K_{w}, K_{l})$ by:
\begin{equation}\label{eq:combine}
    \mathcal{D} = \{ (I_{l}, K_{w}, K_{l}) \mid (K_{w}, K_{l}) \in \mathcal{P}^{db} \cup \mathcal{P}^{sql}\}.
\end{equation}
\subsection{Preference Learning with PLDBF}
Typically, preference fine-tuning is employed subsequent to SFT for further refinement.
In our scenario, the preference learning framework utilizes a direct preference optimization (DPO)~\cite{rafailo2023direct} algorithm. For each preference pair $(I_{l}, K_{w}, K_{l}) \in \mathcal{D}$, the objective function of this process can be formulated as:
\begin{align}
& \mathcal{L}_{\mathrm{PL}}(\pi^{DPO}; \pi^{SFT}) \\
& =  - \mathbb{E}_{\pi}[\log \sigma(\beta R(K_{w}) - \beta R(K_{l}))], \nonumber
\end{align}
specifically, $\mathbb{E}_{\pi}=\mathbb{E}_{(I_{l}, K_{w}, K_{l}) \sim \mathcal{D}}$ and $R(K)=\log (\pi^{DPO}(K | I_{l}) / \pi^{SFT}(K | I_{l}))$.
where $R(\cdot)$ is the reward implicitly defined by the target model $\pi^{DPO}$ and reference model $\pi^{SFT}$.
Then, we obtain the DPO refined model $\pi^{DPO}$, which can generate expert knowledge to assist in accurate database execution and contribute to precise SQL generation.

\subsection{Knowledge-to-SQL with DELLM}
Finally, the user question, the database schema, and the relevant table are given during the testing phase. 
With PL-refined DELLM $\pi^{DPO}$, we collect the generated expert knowledge and combine it with the question, schema, and task instruction to assist an off-the-shelf text-to-SQL model for SQL generation. 
We enumerate each question and the corresponding schema and get the predicted SQL query as the result.

\section{Experiments}
In this section, we will empirically evaluate our proposed Knowledge-to-SQL framework. After introducing the experimental setups, the experimental results are discussed in five parts:
\begin{itemize}
    \item \textit{Main Result:} The purpose of our framework is to assist a text-to-SQL model in generating accurate SQL. We compare the originally predicted SQL with the predicted SQL incorporating the generated expert knowledge; the comparative performance is evaluated on various models or methods on different benchmarks.
    \item \textit{Ablation Study:} We conduct the ablation studies to verify the efficiency and robustness of our proposed framework. By predicting SQL using the knowledge generated by variations of DELLM, we discuss the influence of different modules.
    \item \textit{Comprehensive Evaluation:} We first look deep into how the generated knowledge assists different types (difficulties) of the question. Then, we compare the improvement brought by the generated knowledge and the ground-truth knowledge. Then conclude the comparative results based on the two results above.
    \item \textit{Performance on Partial Training Data:} We discuss the scenario with partial training data and analyze our advantage on practical scenarios with a limited budget.
    \item \textit{Statistical Analysis:} To visualize the influence of DELLM from a data statistical perspective, we calculate the ratio of various influences brought by incorporating the generated knowledge and provide corresponding discussions and analyses.
\end{itemize}

\subsection{Experimental Setup}
\paragraph{Dataset.} Our experiments are conducted on two widely recognized dataset BIRD~\cite{li2023BIRD} and Spider~\cite{yu2018Spider}. BIRD benchmark was released most recently, which annotated high-quality text-to-SQL instances with 95 databases on a large scale. BIRD is a leading benchmark focusing on massive and real database content, first introducing knowledge reasoning between natural language questions and database content. As the evaluation metric, BIRD introduces a new metric verifying the balance of efficiency and accuracy during the execution. The knowledge introduced by BIRD is annotated by human experts who are native speakers of English with degrees above the bachelor’s level and have database-related knowledge. 
Spider benchmark is assessed frequently to evaluate the performance of text-to-SQL across multiple databases, which include 200 distinct databases and 138 domains. 
Since the test set of these two datasets is not publicly available, we evaluate our method's efficacy on the accessible development (dev) set. 

\paragraph{Evaluation Settings.} 
To make a fair comparison, we follow the metric as previous work~\cite{gao2023text,wang2023mac} for evaluation. 
We consider two metrics: 1) Execution Accuracy (\textbf{EX}), which is the ratio of questions that the execution results of the predicted SQL queries match exactly with those of the ground-truth SQL queries, compared to the total number of questions. 
2) Valid Efficiency Score (\textbf{VES}), designed to evaluate the efficiency of SQL generated by text-to-SQL systems. 
The VES metric considers both the efficiency and accuracy of execution results and incorporates running time for a more comprehensive evaluation.
\begin{table*}[!t]
  \centering
\scalebox{0.9}{
  \begin{tabular}{clcccc}
    \toprule
    & \multicolumn{1}{c}{\multirow{2}{*}{\textbf{Models}}} & \multicolumn{2}{c}{\textbf{EX}} & \multicolumn{2}{c}{\textbf{VES}}  \\
    \cmidrule(lr){3-4} \cmidrule(lr){5-6}
    & & {w/o knowledge} & {w/ DELLM} & {w/o knowledge} & {w/ DELLM} \\
    \midrule
    \multirow{7}{*}{\rotatebox[origin=c]{90}{BIRD}} 
    & T5-3B & 10.37 & 16.68 \textcolor{gray}{(+6.31)} & 13.62 & 20.84 \textcolor{gray}{(+7.22)} \\
    \cmidrule{2-6}
    & GPT-3.5-Turbo & 27.64 & 33.31 \textcolor{gray}{(+5.67)}  & 28.64 & 36.12 \textcolor{gray}{(+7.48)} \\
    & GPT-4 & 33.25 & 37.94 \textcolor{gray}{(+4.69)} & 35.92 & 42.15 \textcolor{gray}{(+6.23)} \\
    & Claude-2 & 30.05 & 35.53 \textcolor{gray}{(+5.48)} & 32.97 & 39.71 
    \textcolor{gray}{(+6.74)} \\
    \cmidrule{2-6}
    & GPT-3.5-Turbo + CoT & 27.25 & 32.79 \textcolor{gray}{(+5.54)} & 29.16 & 35.51 \textcolor{gray}{(+6.35)}\\
    & DAIL-SQL + GPT-4 & 40.89 & 45.81 \textcolor{gray}{(+4.92)} & 45.13 & 51.59 \textcolor{gray}{(+6.46)}\\
    & MAC-SQL + GPT-4 & 43.65 & 48.92 \textcolor{gray}{(+5.27)} & 48.07 & 54.78 \textcolor{gray}{(+6.71)} \\
    \midrule
    \addlinespace[4pt]
    \multirow{2}{*}{\rotatebox[origin=c]{90}{Spider}} 
    & GPT-3.5-Turbo & 67.89 & 69.60 \textcolor{gray}{(+1.71)} & 68.33 & 70.16 \textcolor{gray}{(+1.83)} \\
    & GPT-4 & 70.02 & 71.68 \textcolor{gray}{(+1.66)} & 71.03 & 72.82 \textcolor{gray}{(+1.79)} \\
    \addlinespace[3pt]
    \bottomrule
\end{tabular}}
\caption{Experimental results for text-to-SQL on different benchmarks with and without knowledge generated by our proposed DELLM. The number in the bracket denotes the improvement in execution accuracy (EX) and valid efficiency score (VES) brought by DELLM's knowledge compared to the baseline performance without knowledge.}
\label{tab:main-results}
\end{table*}
\paragraph{Implementations.}
We select the widely-used LLaMA-2-13b~\cite{touvron2023llama2} with official configuration as our backbone model for knowledge generation. The learning rate is set as 5e-05, selected from the interval [1e-05, 1e-04]. The semantic similarity calculation of the table reading process is implemented by Faiss\footnote{\href{https://github.com/facebookresearch/faiss/}{https://github.com/facebookresearch/faiss/}}~\cite{douze2024faiss}. The other hyper-parameters are discussed in the Appendix~\ref{sec:appendix}. 
Noting that the Spider dataset does not have annotated knowledge samples, we utilize the human-annotated knowledge in BIRD as ground-truth knowledge to train the model and evaluate the main performance in both the BIRD and Spider datasets. 
The evaluation of the Spider dataset aims to verify our effectiveness in cross-domain databases. The ablation study and further discussions are conducted on the BIRD dataset.
 
\paragraph{Baselines.}
We select the comparative baselines on the BIRD dataset based on the official benchmark page\footnote{\href{https://bird-bench.github.io/}{https://bird-bench.github.io/}}. 
Specifically, to verify the effectiveness of DELLM on the open-source model, we utilize 1) T5-3B~\cite{raffel2020exploring}, using the fine-tuning-based method that incorporates the knowledge as fine-tuning input. Then, for the off-the-shelf models, we utilize 2) GPT-3.5-Turbo and 3) GPT-4~\cite{openai2023gpt}, a widely-used powerful proprietary model with zero-shot text-to-SQL prompt for SQL generation; 4) Claude-2~\cite{claud2report}, another well-recognized proprietary model. For the up-to-date prompt engineering techniques, we compare 5) GPT-3.5-Turbo + CoT~\cite{li2023BIRD}, a simple CoT prompt engineering on text-to-SQL; 6) DAIL-SQL~\cite{gao2023text}, encoding structure knowledge and selects few-shot instances based on similarity matching; 7) MAC-SQL~\cite{wang2023mac}, a novel LLM-based multi-agent collaborative framework designed for the text-to-SQL task. On the Spider dataset, as introduced above, the evaluation is conducted as a cross-domain verification. We utilize proprietary models GPT-3.5-Turbo and GPT-4 for validation.
\subsection{Main Results}
Table~\ref{tab:main-results} gives the experimental results. The number in the brackets represents the improvement of the metric by prompting the LLMs with knowledge generated by DELLM. 

\paragraph{BIRD Results.}
The knowledge generated by DELLM obtained promising results on both metrics in assisting different models/prompting techniques in the text-to-SQL task, which obtained around 5\% improvement on EX and 6\% in VES around all comparative baselines. 
Specifically, 1) The knowledge significantly improves the model's performance with simple prompting. We owe this phenomenon to the PL-refinement based on the database execution and ground-truth SQL contribution feedback, which enable DELLM to generate more helpful knowledge to assist SQL generation. The straightforward prompt challenges the LLMs' capability for question and schema understanding, especially when the question is challenging. DELLM generates the corresponding expert knowledge, reducing the difficulty of understanding, which leads to substantial improvement. 2) When assisting prompting techniques, the knowledge also achieves promising results. The prompting technique surpasses the simple prompt by focusing on providing high-quality few-shot instances or decomposition, which is a various angle from knowledge that assists in enhancing SQL generation. 
This means that DELLM can potentially improve the state-of-the-art performance from a knowledge-assisting angle when facing the scenario without annotated knowledge. Although the well-designed techniques achieve solid performance, there is still space for improvement in generating helpful knowledge.

\paragraph{Spider Results.}
The database scenario of Spider is less challenging. The performance on the same metrics is overall better, as we can see from the results.
Similar to BIRD, 1) The knowledge substantially improves the performance of proprietary models. 2) The improvement brought by the knowledge is significantly lower than the BIRD dataset. We owe this result to the divergent complexity and characteristics of the database structure and the different difficulties of the user question. The question in the Spider dataset may not be as challenging as in the BIRD. As the number of questions that need to be assisted decreases, the improvement declines respectively.

\subsection{Ablation Study}
Table~\ref{tab:ablation} presents the results of the ablation study for our framework in the BIRD dev set with proprietary LLM GPT-4. The table lists different variations of DELLM, including excluding database execution feedback (\textit{db feedback}) or ground-truth SQL contributing feedback (\textit{sql feedback}) and also removing the \textit{table reading} process. The column presents the evaluation metrics.

\begin{table}[t]
\centering
\scalebox{0.9}{
\begin{tabular}{lcc}
\hline
\textbf{Models} & \textbf{EX} & \textbf{VES} \\ 
\hline
GPT-4 + DELLM & 37.94 & 42.15  \\
w/o \textit{table reading} & 37.23 \textcolor{gray}{(-0.71)} & 41.30 \textcolor{gray}{(-0.85)}\\
w/o \textit{db feedback} & 36.25 \textcolor{gray}{(-1.69)} & 40.46 \textcolor{gray}{(-2.07)}\\
w/o \textit{sql feedback} & 36.91 \textcolor{gray}{(-1.03)} & 41.12 \textcolor{gray}{(-1.55)}\\
\hline
\end{tabular}
}
\caption{Ablation study on variations of DELLM.}
\label{tab:ablation}
\end{table}

As we can see in Table~\ref{tab:ablation}, removing any components leads to a performance decrease on both metrics. The performance declined the most when excluding the database feedback, indicating that assisting the predicted SQL in database execution is the priority in generating helpful knowledge. Overall, the ablation study demonstrates that each component of our method plays an irreplaceable role in achieving good performance, as their removal decreased the performance as intuitively expected.

We also discuss the choice of PL algorithm. Apart from the utilized direct preference optimization (DPO)~\cite{rafailo2023direct} in our framework, we evaluate a variation of DELLM, which is PL-refined by proximal policy optimization (PPO)~\cite{schulman2017proximal}.

\begin{table}[t]
\centering
\setlength\tabcolsep{8pt}
\scalebox{0.9}{
\begin{tabular}{lcc}
\hline
\textbf{Models} & \textbf{EX} & \textbf{VES} \\ 
\hline
GPT-4 & 33.25 $\pm$ 0.61 & 35.92 $\pm$ 1.03  \\
+ $\text{DELLM}_{PPO}$ & 35.28 $\pm$ 1.18 & 40.03 $\pm$ 1.81  \\
+ $\text{DELLM}_{DPO}$ & 37.94 $\pm$ 0.57 & 42.15 $\pm$ 0.95  \\
\hline
\end{tabular}
}
\caption{Ablation study on the utilized PL algorithms.}
\label{tab:ablation-PL}
\end{table}

Given the results in Table~\ref{tab:ablation-PL}, we can conclude: 1) PPO works for DELLM, which can also generate helpful knowledge to improve the performance of the text-to-SQL model.
2) DPO outperforms PPO, one potential reason is that PPO relies on a reward model to give numeric rewards for each generated knowledge. Since the knowledge generation task is non-deterministic, even if the provided knowledge is correct, the downstream SQL generation may be wrong. In this case, the DPO algorithm with implicit rewards performs better.

\subsection{Comprehensive Evaluation}
To reach a comprehensive conclusion, we discuss SQL generation at different difficulty levels and compare the knowledge generated by DELLM to the ground-truth knowledge. Table~\ref{tab:diff} shows the results on the BIRD dev set, decomposing different difficulty levels of the user question. In the table, D and E represent the knowledge generated by our proposed DELLM (D) and the ground-truth knowledge annotated by human experts (E).
\begin{table}[!t]
  \centering
  \setlength\tabcolsep{4pt}
  \scalebox{0.9}{
  \begin{tabular}{lcccc}
    \toprule
    \textbf{Model} & \textbf{Simp.} & \textbf{Mod.} & \textbf{Chall.} & \textbf{All}\\
    \midrule
    GPT-3.5-Turbo & 35.58 & 14.60 & 17.61 & 27.64  \\
    \rowcolor{lightgray}
    GPT-3.5-Turbo + D & 43.09 & 18.30 & 17.61 & 33.31  \\
    \rowcolor{lightgreen}
    GPT-3.5-Turbo + E & 50.27 & 31.81 & 20.42 & 41.98 \\
    \bottomrule
    \midrule
    GPT-4 & 41.05 & 21.13 & 21.13 & 33.25  \\
    \rowcolor{lightgray}
    GPT-4 + D & 47.16 & 24.18 & 21.83 & 37.94 \\ 
    \rowcolor{lightgreen}
    GPT-4 + E & 54.01 & 36.38 & 31.69 & 46.67 \\
    \bottomrule
  \end{tabular}
  }
  \caption{The execution accuracy (EX) for questions with different difficulty. Simp. denotes Simple, Mod. denotes Moderata, and Chall denotes Challenging.}
  \label{tab:diff}
\end{table}

From the results, 1) There is a gap between the DELLM-generated knowledge and the human expert-annotated one on all difficulty levels. The human-annotated knowledge surpasses 8.67\% and 8.73\% to DELLM on execution accuracy, leaving a great gap for improvement. 2) The generated knowledge mainly works on simple and moderate questions. One reason is that LLMs have difficulty understanding the challenging question and schema, making it hard to utilize corresponding generated knowledge. Another reason is that challenging questions only occupy a small portion of the annotations (around 9\%), leading to a data imbalance during the training process.

\subsection{Performance on Partial Training Data}
We select partial annotated knowledge in the BIRD training set to train the DELLM, then discuss the performance of that scenario. We conducted the experiments using GPT-4, and the results are shown as the improvement with the generated knowledge by DELLM on different ratios of training data.

\begin{figure}[!t]
  \centering
  \begin{subfigure}{.24\textwidth}
    \centering
    \includegraphics[width=.9\linewidth]{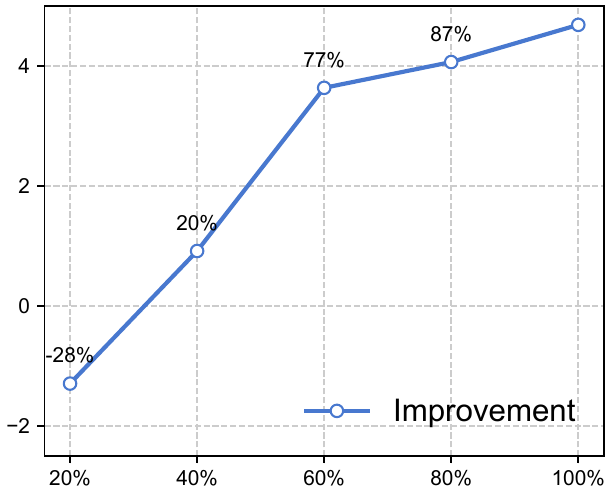}
    \caption{Execution Accuracy}
    \label{fig:ex}
  \end{subfigure}%
  \begin{subfigure}{.24\textwidth}
    \centering
    \includegraphics[width=.9\linewidth]{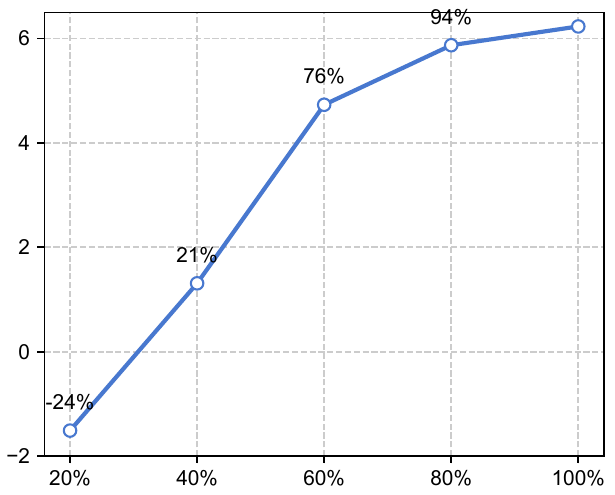} 
    \caption{Valid Efficiency Score}
    \label{fig:ves}
  \end{subfigure}
  \caption{Improvement to GPT-4 on different metrics with DELLM on different ratios of training data.}
  \label{fig:partial}
\end{figure}

As shown in Figure~\ref{fig:partial}, where the x-axis represents the ratio of training data, the y-axis is the value of the metric's improvement. In practice, with a limited budget, we can hire human experts to annotate about 60\% of training data to obtain around 77\% performance for expert knowledge generation of DELLM.

\subsection{Statistical Analysis}
We identify four types of influences that the DELLM's knowledge may bring: 1) Assistance, where the original predicted SQL is not correct but becomes correct with the help of the knowledge. 2) Misleading, where the original predicted SQL is correct, but becomes incorrect due to the influence of the knowledge. 3) Inoperative, where the original incorrect SQL remains incorrect despite the knowledge. 4) Sustainable, where the originally predicted SQL is correct and remains correct with the knowledge. The statistics are obtained based on the experiment result using GPT-4 on the BIRD dev set.
\begin{figure}
    \centering
    \includegraphics[width=0.9\linewidth]{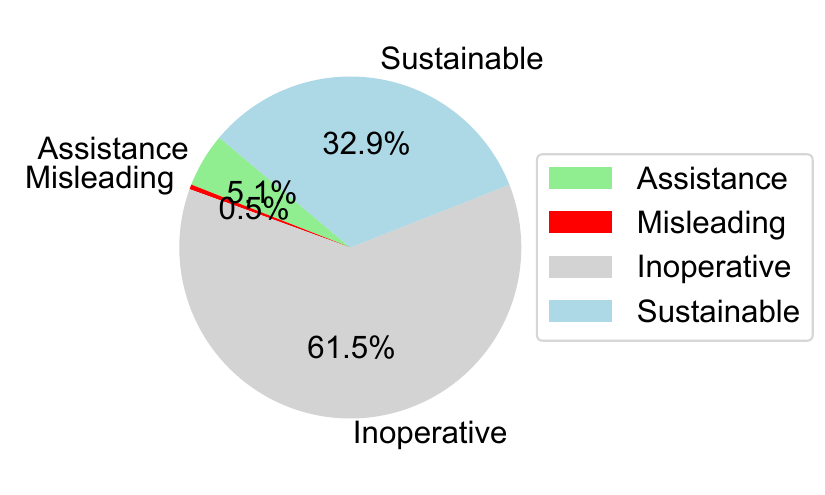}
    \caption{Different influences of DELLM bring on GPT-4's performance on the BIRD dev set.}
    \label{fig:stat}
\end{figure}
From Figure~\ref{fig:stat}, the DELLM brings around 5\% assistance, and there is potentially 0.5\% misleading of its generated knowledge. The inoperative knowledge leaves about 60\% in DELLM and also around 53\% in the expert annotation. One potential optimization to text-to-SQL is improving the effectiveness of incorporating expert knowledge.

\section{Conclusion}
In conclusion, our study highlights the significance of knowledge generation in enhancing the performance of LLMs for text-to-SQL tasks. By proposing a novel framework that leverages database content, execution feedback, and comparisons with ground-truth SQL, we address the challenges of bridging the knowledge gap between user questions and database schema. Through extensive experiments on BIRD and Spider datasets, our approach brings substantial improvements in execution accuracy and valid efficiency scores for models like GPT-4, underscoring its efficacy in advancing text-to-SQL research and fostering innovations in natural language processing and data mining.

\section{Limitations}
Two primary limitations are acknowledged in this study. Firstly, the performance of DELLM is mostly evaluated using proprietary models. However, the in-context learning and natural language understanding capability of open-source medium-scale local LLMs still have a large gap compared to proprietary LLMs. Although with the assistance of generated knowledge, their performance significantly improved, the overall performance is not good enough for practical use. This may limit the application of DELLM in offline deployment situations. Secondly, the generalizability of the proposed framework to real-world scenarios remains to be determined, as it has only been evaluated on standard benchmarks. The complexity, diversity, and potential noise in real-world database environments calls for further verification to confirm the framework's effectiveness in practical applications.

\section*{Ethics Statement}
We confirm that we have fully complied with the ACL Ethics Policy in this study. All the datasets are publicly available and have been extensively used in research related to text-to-SQL.

\section*{Acknowledgements}
This work was supported in part by the Innovation and Technology Commission of HKSAR and the Ministry of Science and Technology of China under the Innovation and Technology Fund - Mainland-Hong Kong Joint Funding Scheme (MHP/012/21), and in part by the Innovation and Technology Commission of HKSAR under the Research Talent Hub for ITF projects (Pih/054/23).

\nocite{*}
\bibliography{custom}
\bibliographystyle{acl_natbib}

\newpage

\appendix

\section{Implementations Details.}\label{sec:appendix}
We discuss the implementation details of our proposed framework. We used parameter-efficient fine-tuning (PEFT) in the training stage to train our models. Specifically, in each stage (supervised fine-tuning and DPO preference learning), we utilize low-rank adaptation (LoRA)~\cite{hu2021lora} as PEFT method, the trainable parameters occupy 0.0622\% of full parameters. All the experiments are conducted on four NVIDIA A800-SXM4-80GB GPUs, and the transformers package version is 4.36.2. The details of training and hyper-parameters are listed as follows. 
\subsection{Supervised Fine-tuning}
As previously introduced, the backbone model of DELLM is selected as LLaMA-2-13b with official configuration\footnote{\href{https://huggingface.co/meta-llama}{https://huggingface.co/meta-llama}}; the training details are given in Table~\ref{tab:sft-appendix}. The model's architecture is identical to the official provided in Huggingface.
\begin{table}[h]
\centering
\setlength\tabcolsep{12pt}
\begin{tabular}{cc}
\hline
Hyper-parameters & Value \\
\hline
data type & fp16 \\
learning rate & 5e-05 \\
number of epochs & 3 \\
number of batch size & 32 \\ 
gradient accumulation steps & 4 \\
\hline
\end{tabular}
\caption{Hyper-parameters of SFT.}
\label{tab:sft-appendix}
\end{table}\\

\subsection{DPO Training}
The hyper-parameters of DPO are similar to the previous stage, the details are shown in Table~\ref{tab:dpo-appendix}.
\begin{table}[h]
\setlength\tabcolsep{12pt}
\centering
\begin{tabular}{cc}
\hline
Hyper-parameters & Value \\
\hline
data Type & fp16 \\
learning rate & 1e-05 \\
number of epochs & 1 \\
number of batch size & 16 \\ 
gradient accumulation steps & 4 \\
\hline
\end{tabular}
\caption{Hyper-parameters of the DPO.}
\label{tab:dpo-appendix}
\end{table}

\subsection{Generation Configuration}
In each generation of parts of our framework during training and testing, the configuration is identical. The details are listed in Table~\ref{tab:gen-appendix}.
\begin{table}[h]
\setlength\tabcolsep{20pt}
\centering
\begin{tabular}{cc}
\hline
Configuration & Value \\
\hline
top p & 0.9 \\
do sample & True \\
temperature & 0.6 \\
max token length & 4096 \\
predict with generate & True \\
\hline
\end{tabular}
\caption{Generation configuration}
\label{tab:gen-appendix}
\end{table}\\

\subsection{Versions of Proprietary LLMs}
The version of ChatGPT in this work is GPT-3.5-Turbo-1106, and the version of GPT-4 is GPT-4-0613.

\end{document}